\documentclass{article}

\usepackage[nonatbib, preprint]{neurips_2023}

\usepackage{graphicx}
\usepackage{amsmath}
\usepackage{amssymb}
\usepackage{booktabs}

\usepackage[utf8]{inputenc} 
\usepackage[T1]{fontenc}    
\usepackage{hyperref}       
\usepackage{url}            
\usepackage{booktabs}       
\usepackage{amsfonts}       
\usepackage{nicefrac}       
\usepackage{microtype}      
\usepackage{xcolor}         

\title{LLM2Loss: Leveraging Language Models for Explainable Model Diagnostics}

\author{%
  Shervin Ardeshir\\
Netflix\\
{\tt\small shervina@netflix.com} \\
}

\begin{document}

\maketitle

\begin{abstract}
Trained on a vast amount of data, Large Language models (LLMs) have achieved unprecedented success and generalization in modeling fairly complex textual inputs in the abstract space, making them powerful tools for zero-shot learning. Such capability is extended to other modalities such as the visual domain using cross-modal foundation models such as CLIP\cite{clip}, and as a result, \textit{semantically meaningful representation} are extractable from visual inputs. 

In this work, we leverage this capability and propose an approach that can provide semantic insights into a model's patterns of successes, failures, and biases. Given a black box model, its training data, and task definition, we first calculate its task-related loss for each data point. We then extract a \textit{semantically meaningful representation} for each training data point (such as CLIP embeddings from its visual encoder) and train a lightweight model which maps this \textit{semantically meaningful representation} of a data point to its task loss. We show that an ensemble of such lightweight models can be used to generate insights on the performance of the black-box model, in terms of identifying its patterns of failures and biases.\footnote{Disclaimer: we used Microsoft Bing as an assistive tool in writing this manuscript.}
\end{abstract}

\section{Introduction}
\label{sec:intro}

\begin{figure}[h]
    \centering
    \includegraphics[width=0.55\textwidth]{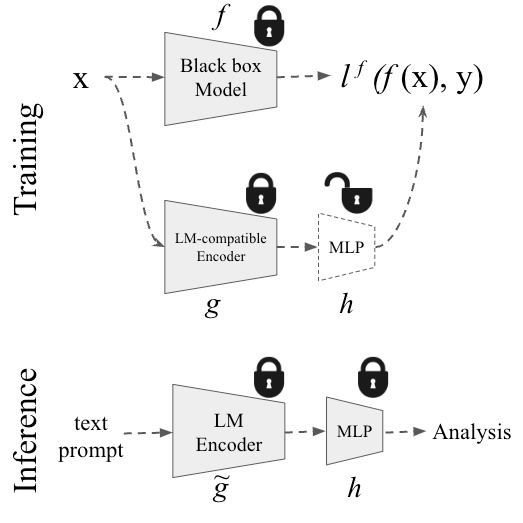}
    \caption{An overview of the proposed approach. Given a black-box model ($f$) trained on a task ($f(x)\rightarrow y$), we aim to estimate the function $h(g(.))$ which estimates its loss on task $f$, i.e. $l^f(f(x), y)$, such that $g(x)$ is an embedding compatible with a language model. As a result, at inference time, we can feed textual descriptions and derive insights into the model's performance on such data points.} 
    \label{fig:teaser}
\end{figure}

Large language models learn to generate natural language texts based on a large corpus of textual data. They have shown promise in modeling the semantic space, which is the representation of meaning and concepts in natural language. Using natural language prompts, which are inputs that guide the model to complete a certain task or generate a certain output, large language models can perform various natural language processing tasks without requiring any task-specific training data or labels. This is also known as zero-shot learning, where the model leverages its general knowledge and linguistic skills to infer the desired output from the prompt. Built on-top-of, or in-conjunction-with language models, vision-to-text models such as CLIP are able to map visual space to the same semantic space as natural language by objective functions such as cross-modal contrastive\cite{clip}. This means that such models learn to align visual and language representations by maximizing their similarity for pairs of images and texts that are semantically related and minimizing their similarity for pairs that are not \cite{clip}. By doing so, vision-to-text models can achieve a better understanding of both modalities and perform various tasks such as image captioning, image retrieval, visual question answering, etc. As a result, such language based models \cite{gpt3, clip} are capable of producing highly structured embeddings spaces, which also are inherently semantic, and thus, human interpretable/understandable. Here we aim to leverage this characteristic to derive some insight from the performance of a black-box machine learning model, in terms of understanding it's potential areas of strengths, failures, and biases post-hoc and in absence of any given ground-truth semantic categories. 

Given a black box model and its training data, we first calculate its loss for each data point using a single forward pass. We then extract a semantically meaningful, and Language-model-compatible representation for each training datapoint (such as CLIP). We then train a lightweight diagnosis model which maps the semantically meaningful representation of a datapoint to its task loss. We show that such a model can provide semantic insight in terms of identifying patterns of failures and biases in the performance of models.

\section{Related Work}
Language models have shown promise in capturing fairly complicates structures within the semantic space \cite{gpt3, touvron2023llama, chowdhery2022palm}.
On the other hand, understanding of model robustness and estimating it across an abstract and contineous embeddinng space has been explored in \cite{Ardeshir_2022_CVPR}. Here we aim to add the capabilities of LLMs to that domain and allow for estimation of robustness in a way that is semantically interpretable. This is a fairly new area, and relevant to this line of work may be semantic adversarial approaches \cite{joshi2019semantic, qiu2020semanticadv, sauer2021counterfactual} which rely on generative models in order to generate adversarial examples, and \cite{harkonen2020ganspace} where interpretable controls are discovered for generative models. However, our work explores a complementary direction, and simpler approach to understanding a models failures and biases without the need for generating any artificial input data. Another line of related work which would be relevant to ours is that of \cite{berg2022prompt, ramaswamy2021fair, qi2023improving} where the authors aim at using latent spaces to improve model robustness. 

Zeros shot capability of CLIP \cite{clip} embeddings have shown promise in many applications such as manipulation of NERFs \cite{wang2022clip}, avatars \cite{hong2022avatarclip}, shape generation \cite{sanghi2022clip}, and debiasing \cite{berg2022prompt}.

\section{Approach}
An overview of the approach is shown in Figure \ref{fig:teaser}. We assume that we are given a black-box model trained to map input $\mathbf{X}=\{x_i\}_{i=1,..., N}$ to output $\mathbf{y}=\{y_i\}_{i=1,..., N}$ via function $f: \mathbf{x} \rightarrow \mathbf{y}$ through minimization of loss function $l^f(f(x), y)$. In our experiments, we set $f$ to be a face recognition model trained on the CelebA dataset\cite{celeba}. We also assume that the format of $x_i$s is compatible with some Language compatible model $g$. By language compatible, we mean that there is a foundation language-based model that can map this input to a common space with language models. Given that the input data for the face recognition model is visual, in our experiments we used CLIP visual encoder as the language-compatible encoder $g$, and CLIP textual encoder as $\Tilde{g}$.


\subsection{Training}
Our goal is to train model $h$, such that it can predict model $f$'s loss for datapoint $x_i$, i.e. $l^f(f(x_i), y_i)$, solely from the content of its input $x_i$. Given that loss values are often continuous, we experiment with two variations of this mapping functions, namely, a regression model, and a ranking model. We trained both regressor and ranker alternatives in an ensemble setup to be able to have uncertainty on each of the generated insights. We trained our ensemble models in a very simple setup, and by training 5 independent runs with randomized initializations.
\subsubsection{$h$ as a Loss Regressor}
\label{sec:h_regressor}

The regression model is trained using MSE loss and defined as:   
\begin{equation}
l_{regr}^h = MSE(h(g(x_i)),  l^f(f(x_i), y_i)),
\end{equation}
in other words, $h$ estimates the absolute value of the task loss. Intuitively, this would estimate how much error model $f$ is expected to have for a given input. 

\subsubsection{$h$ as a Loss Ranker}
\label{sec:h_ranker}

An alternative is to train $h$ using a ranking loss. In this setup we sample pairs of datapoints $(x_i, x_j)$, and train $h$ to be able to predict which datapoint leads to a higher loss through $f$. We train this model using a binary-crossenropy loss defined as:
\begin{equation}
l_{rank}^h = BCE(\Delta h(g(x)),  \Delta l^f(f(x), y) > 0),
\end{equation}

in which: 
\begin{equation}
\Delta h(g(x)) = h(g(x_i)) - h(g(x_j)), 
\end{equation}
and: 
\begin{equation}
\Delta l^f(f(x), y) = l^f(f(x_i), y_i) - l^f(f(x_j), y_j).
\end{equation}

Intuitively, this model predicts if model $f$ does better for datapoint $x_j$ compared to datapoint $x_i$.
In our setup, the face recognition model was trained using a multi-class classification setup (predicting identity as a class label from face image content), and therefore the per-datapoint loss $l^f(f(x_i), y_i)$ is the output of a categorical cross-entropy loss function, and therefore a continuous value.


\subsubsection{$h$ as a Classifier}
\label{sec:h_classifier}
Another aspect of trying to understand a model's strengths and weaknesses is to understand whether the model has "seen" a specific type of input during training or not. In other words, is input $\mathbf{x}$ \textit{in-distribution} when compared to it's training dataset. A datapoint being seen or unseen can have a significant impact on a models predictions for this datapoint being meaningful. More often than not, such predictions are not meaningful. As a result, we also believe that the estimated values for the losses may or may not be valid for a datapoint depending on such phenomenon. Thus, we experiment with equipping our diagnostics to a secondary output which indicates if a datapoint is in or out of distribution. We model such an output as a binary classification problem: in-distribution (\textit{i.d.}) vs out-of-distribution (\textit{o.o.d.}), or seen vs unseen. Given that each training datapoint results in a 512 dimesnional CLIP embedding, we generate random 512 dimensional vectors as out-of-distribution areas of the embedding space. We then train a binary classifier on in-distribution vs out-of-distribution embeddings. Our assumption is that we only have access to the training datapoints of the task of interest, and thus we can only use random negatives. Our experiments (included in Section \ref{sec:qualitative_seen_unseen} and Figure \ref{fig:qualitative_seen_unseen}) indicate that such a simple approach achieves meaningful results in terms of allowing a user to understand what type of datapoints have been seen by a model. Also, our experiments suggest that the loss-regressor (explained in Section \ref{sec:h_regressor}) already has this capability and is strongly capable of OOD detection, which in fact suggests that there may not be a need for a separate OOD detection component. However, we still experimented with this variation to provide a good point of comparison to the OOD detection capabilities of the loss-regressor and loss-ranker models. 

\subsection{Inference}
As shown at the bottom row of Figure \ref{fig:teaser}, at inference time, we aim to provide some insights on a given textual input $x_{text}$. Intuitively, given textual input $x_{text}$, $-h(\Tilde{g}(x_{text}))$, will estimate, how good the model's performance is on $x_{text}$ type datapoints. As an example, we can feed textual inputs such as \textit{$x_{text}$="face image of a long-haired smiling male wearing glasses"} and get an estimated loss for that. Assuming generalization, this is an estimate of how good the black-box model $f$ is on its own task, on inputs that resemble $x_{text}$. Given that absolute values of a loss are often hard to interpret, a better way would be to feed two different textual inputs and compare their estimated performance. As an example feeding \textit{"face of a smiling long-haired male"}, and \textit{"face of a smiling long-haired female"}, and observe if the model's performance has a bias towards the former or the latter. In the OOD classifier variation of $h$, the same analysis would tell us how familiar a model is with the inputs that resemble $x_{text}$, in other words, if semantic concept $x_{text}$ is in-distribution with respect to $f$'s training data.

\section{Experiments}
In our experiments, we used celebA dataset\cite{celeba} and used a trained face-recognition model as $f$. We use categorical cross-entropy loss of the face recognition model as $l^f$, and the target for $h$, and experimented with simple linear and (2-layer) non-linear MLPs for the architecture of $h$, for both variations of regressor and ranker. Given that we used CLIP \cite{clip} encoders as $g$ and $\Tilde{g}$, $g(x_i)$ is a 512-dimensional vector resulting from the image encoder of CLIP. In inference mode, we use the text encoder of CLIP as $\Tilde{g}$. The main application of our proposed approach in realistic scenarios would be better demonstrated using qualitative examples. Thus we demonstrate a few examples of how we can use the trained model \textit{h} to derive insights on model bias and failures. We then aim to measure the correlation between our generated insights to their true values based on attributes available for the celebA \cite{celeba} dataset. 

\subsection{Loss/Prediction Validity: Explainable/Semantic OOD Detection}

As mentioned earlier, an important aspect of understanding a models strengths and weaknesses is to understand on which datapoints a models predictions are valid. In other words, whether a datapoint is in-distribution or out-of-distribution for a model. The classifier variation of $h$ is explicitly trained to do so. In Section \ref{sec:qualitative_seen_unseen} we visualize how insights about the validity of predictions of a model could be derived using this variation pf $h$, and in Section \ref{sec:quantitative_ood} we quantify the performance of different variations of $h$ in terms of their capability in performing OOD detection in the semantic space (more details in Section \ref{sec:quantitative_ood}).

\subsubsection{Qualitative}
\label{sec:qualitative_seen_unseen}

\begin{figure}[h]
    \centering
    \includegraphics[width=0.48\textwidth]{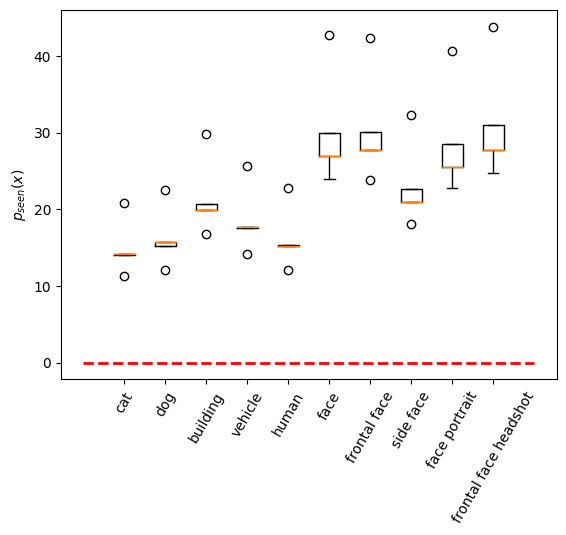}
        \includegraphics[width=0.48\textwidth]{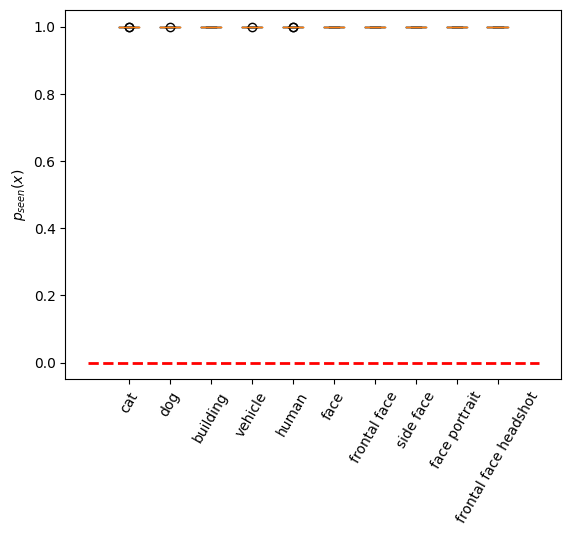}
    \caption{Explainable OOD detection. Right: post-sigmoid. Left: pre-sigmoid. It can be observed that phrases describing face data are correctly identified as better descriptions of the data that the model has seen.} 
    \label{fig:qualitative_seen_unseen}
\end{figure}

Here we visualize how insights about the validity of predictions of a model could be derived using the classifier variation of $h$. Figure \ref{fig:qualitative_seen_unseen} shows the output of prediction of $h_{classifier}$ on a set of textual inputs. Right shows the output of the classifier post the Sigmoid activation, and left shows the output prior to it, to signify the differences. As it can be observed, textual inputs such as face, frontal-face, and frontal face headshot are predicted to be more in-distribution compared to concepts such as cat, dog, building. Given that model $f$ in our experiments is a face recognition model trained on celebA it's only been exposed to face images. Thus such textual inputs are correctly predicted to be more in-distribution / familiar to the model. Interestingly, even within different variations of textual input containing the word face, the ones containing the phrase "frontal face" has much higher scores than the one containing "side face". Knowing that celebA primarily contains frontal face headshots, our anecdotal qualitative examples seem to capture the seen vs unseen domains fairly meaningfully. We argue that the reason why the post Sigmoid probabilities are all close to 1 is that in our very simple training setup ($x_{tr}$ vs. $x_{rand}$), the negative samples are randomly generated. As a result, the classifier can draw the boundary between random vectors and a manifold containing all natural images, resulting in all natural textual inputs being classified as seen/in-distribution. We argue that this is something that can easily improved once out-of-distribution samples come from natural images/inputs and not randomly generated. However, our experiments indicate that even without that and with this extremely simple way of randomly generated negatives, the model meaningfully learns that faces are more familiar to the model compared to concepts such as cars,dog, etc. We later quantify this capability of $h_{classifier}$ and compare it to  $h_{regressor}$ and $h_{ranker}$, in section \ref{sec:quantitative_ood}.

\subsubsection{Quantitative}
\label{sec:quantitative_ood}

Table \ref{tab:ood_quantitative} shows the performance of our different variations of $h$ on semantic OOD detection task. In this task, we feed semantic inputs describing the seen dataset (such as "face", "portrait", "human face", etc) in addition to out-of-distribution data. We then feed class labels of some out-of-distribution / unseen datasets as $x_{text}$ to the inference pipeline $h(\hat{g}(x_{text}))$. We then evaluate whether the in-distribution class labels are retrieved correctly and with a higher score, \textit{i.e.} measure the AUROC of $h(\hat{g}(x_{text}))$. Rows 2,3, and 4 of Table \ref{tab:ood_quantitative} reflects that metric for classes of CIFAR10 \cite{cifar}, and super-classes and classes of CIFAR100 \cite{cifar}. It can be observed that all models perform better than chance, and interestingly, the loss-regressor variation (\textit{i.e.} $h_{reg}$) performs better than the classification variation (\textit{i.e} $h_{class}$) which is directly trained to perform this task. As a result, the loss regressor model in its ensemble form performs very reliably not only in terms of estimating the loss (as it's directly trained on it), but also the validity of it.      

\begin{table}
\small
\begin{center}
\begin{tabular}{|c|c|c|c|c|c|c|}
\hline
OOD dataset/model& $h_{rank.}$ & $h^{ens-mean}_{rank.}$ & $h_{reg.}$ & $h^{ens-mean}_{reg.}$ & $h_{class.}$ & $h^{ens-mean}_{reg.}$ \\\hline
random & $0.670 \pm 0.217$ & 0.686 & $1.0 \pm 0.0$ & 1.0 & $1.0 \pm 0.0$ & 1.0 \\\hline
CIFAR10-classes & $0.576 \pm 0.049$ & 0.62& $0.919 \pm 0.047$ & 1.0 & $0.848 \pm 0.009$ & 0.840 \\\hline
CIFAR100-super-classes & $0.555 \pm 0.058$ & 0.578 & $0.951 \pm 0.036$ & 0.989 & $0.903 \pm 0.012$ & 0.915 \\\hline
CIFAR100-classes & $0.700 \pm 0.101$ & 0.722 & $0.951 \pm 0.016$ & 0.978 &
$0.878 \pm 0.005$ & 0.88 \\\hline
\end{tabular}
\caption{Loss-validity, \textit{i.e.} explainable OOD Detection. In-distribution dataset is always celebA. Left column specified classes from what out-of-distribution dataset were used in this evaluation.}
\label{tab:ood_quantitative}
\end{center}
\end{table}

\subsection{Loss Estimation: }
\label{sec:loss_estimation}
The regressor and ranker models aim directly at estimating the absolute value and ranking of the model's loss given input data. We acknowledge that it would be hard to derive any insight from an absolute itself, however, comparing performance of a model across two (or more) semantic concepts could be eye opening in terms of model's expected areas of strength and weakness. In this section, we first go over qualitative examples of how insights could be extracted by performing such comparisons in Section. \ref{sec:loss_estimation_qualitative}. We then provide quantitative results in terms of how successful each of our three variations are in terms of achieving such ranking in Section \ref{sec:loss_estimation_quantitative}.   

\subsubsection{Qualitative}
\label{sec:loss_estimation_qualitative}
\begin{figure}[h]
    \centering
    \includegraphics[width=0.4\textwidth]{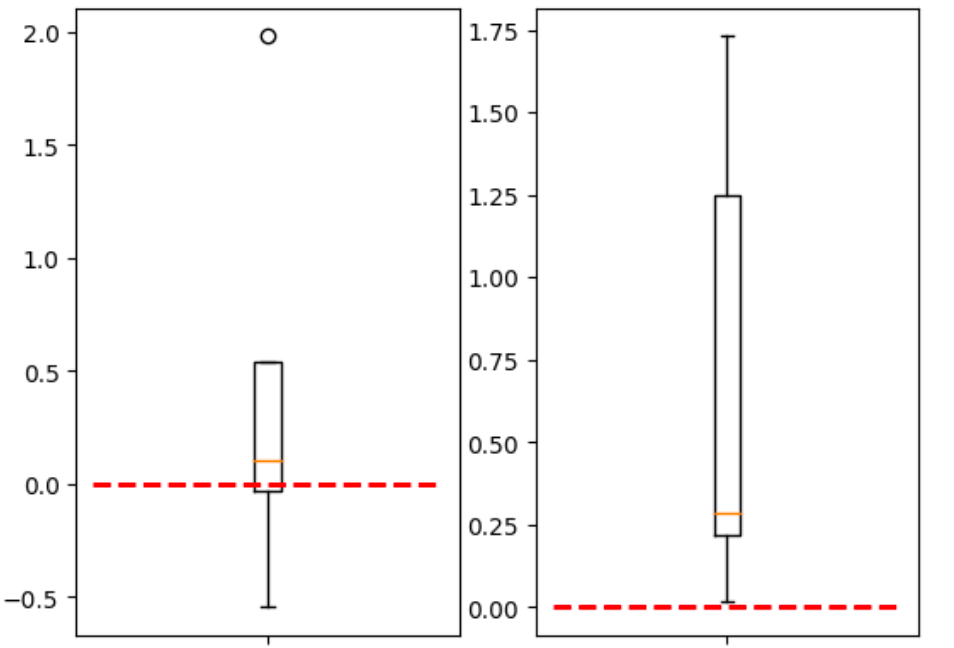}
    \caption{Two examples of pairs of textual inputs fed to the model for bias estimation. Right is the results of feeding the pair $x_1$="face of an old male", and $x_2$="face of a young female". And left is the results of feeding $x_1$="smiling long-haired person wearing glasses", and $x_2$="non-smiling long-haired person wearing glasses". It can be observed that the model has a more confident prediction about the bias across $x_1$="face of an old male" vs $x_2$="face of a young female". The results on the right shows that the model confidently thinks that $f$ has a lower loss for $x_2$, and therefore it's performance is biased towards it.} 
    \label{fig:qualitative}
\end{figure}

Here we show examples of how our approach at inference mode could be used to generate explanations on the potential failure areas of the model, and potentially uncover bias in the model's performance across pairs of input types. We visualize two examples of such insights in Figure \ref{fig:qualitative}. On both examples we are visualizing $h(\Tilde{g}(x_1)) - h(\Tilde{g}(x_2))$ where $x_1$ and $x_2$ are two text inputs across which we would like to estimate model's performance disparity. Right shows the results of feeding the two following textual inputs $x_1$=" face of an old male", and $x_2$=" face of a young female". The figure shows the quadrants of probabilities that loss is higher for $x_1$ compared to $x_2$. It can be observed that the model thinks model $f$ would do better on inputs resembling "face of a young female" compared to inputs resembling "face of an old male", as the loss is predicted to be lower for "face of a young female" compared to "face of an old male". Given that we trained ensembles of models, each individual model will have a point-estimate prediction on the estimated-loss difference, leading to a distribution of estimations from the ensemble. This allows us to measure the likelihood of $h(\Tilde{g}(x_1)) - h(\Tilde{g}(x_2))$ to be above (or below) zero, indicating that the model is biased towards $x_2$ compared to $x_1$ (or vice-versa). The bars show the quadrants of the distribution and as it can be observed the whole distribution is above zero indicating a fairly confident estimation. On the other hand, the example on the left-hand side indicates that for textual inputs $x_1$=" smiling long-haired person wearing glasses", and $x_2$="non-smiling long-haired person wearing glasses" the prediction is less certain. In the following section, we aim at quantifying this capability.  

\subsubsection{Quantitative}
\label{sec:loss_estimation_quantitative}
\textbf{Ranking datapoints based on their loss:}
In this experiment we pass each of the images in celebA test dataset through $h(g(x))$ to get its estimated loss. We then pass the same images through $f$ and measure their actual losses $l(f(x), y)$. We then evaluate how well the two sets correlate. We evaluate this for each type of $h$ (namely ranker, regressor, and classifier), and report the results in Table \ref{tab:loss_ranking_all}. Given that we trained 5 models for each setup, we report the results in ensemble mode (using mean and median aggregation of the scores). We report the performance of the models individually in the supplementary material. It can be observed that all models positively correlate, However, when it comes to the ranking correlation metrics, the ranking model strongly outperforms the other two variations, which is expected as it is trained for this objective. Interestingly, the classifier seems to consistently perform better than chance even though it's objective does not have any notion of absolute or relative value of loss for a datapoint. This suggests the relativity of datapoints being in-distribution. In other words, the datapoints that are \textit{more in-distribution} (and thus probably over represented), tend to have lower losses. This is consistent with the observation in Section \ref{sec:quantitative_ood} where we observed the ranker and regressor having great performance in terms of OOD detection.

\begin{table}
\begin{center}
\begin{tabular}{|c|c|c|c|}
\hline
& Pearson & Spearman & Kendall Tau\\\hline
$h^{ens}_{class}$ & (0.01463, 0.086) & (0.13181, 3.850e-54) & (0.08958, 1.499e-53)\\\hline
$h^{ens}_{reg}$ & (0.06851, 9.97e-16) & (0.27099, 3.75e-229) & (0.18626, 3.76e-225)\\\hline
$h^{ens}_{rank}$ & \textbf{(0.13288, 5.33e-55)} & \textbf{(0.56047, 0.0)} & \textbf{(0.39860, 0.0)}\\\hline
\end{tabular}
\caption{Evaluating performance of the three variations of $h$ in terms of ranking test data-points based on their estimated loss \textit{i.e.} $h(g(x)$. The correlation is evaluated between the predicted losses and the actual losses  $l^f(f(x), y)$ calculated based on model $f$ and ground-truth label $y$.}
\label{tab:loss_ranking_all}
\end{center}
\end{table}

\textbf{Ranking categories based on model's predicted bias:} Given that celebA\cite{celeba} dataset provides textual attributes for each of the data points, we measure the ground-truth bias of model f across those textual attributes. To do so we measure AUROC of the face recognition model for each binary of a semantic attribute. As an example we measure mAUROC for all images with attribute "Young=True", and for all images with attribute "Young=False". We then measure the bias of the model across the two by measuring the Rawlsian \cite{rawls2001justice} max min of the two values. These values for each attribute are reported in the first column of Table \ref{tab:per_attr_results} under GT-True AUROC, GT-False AUROC and GT-RMM (Ground-truth Rawlsian max-min). The higher the Rawlsian Max-Min the higher the disparity of performance across the two categories. We then predict the model's bias using $h(\hat{g}(x_{text}))$. To do so we calculate the Rawlsian max-min between the following two values: $h(\hat{g}(x_{text}="Young"))$, $h(\hat{g}(x_{text}="Not Young"))$. We simply infer on the attribute and itself appended to "Not". We report this estimated Rawlsian Max-Min in column CLIP-RMM of Table \ref{tab:per_attr_results}. Finally, we measure how our predicted bias CLIP-RMM is correlated to the ground-truth bias GT-RMM. In Table \ref{tab:correlations} we report the correlations in terms of Pearson, Spearman, and Kendall-tau ranking correlation. It can be observed that for all three, the results positively correlate with statistically significant p-values. Similar to the per datapoint loss-based ranking task, the ranker variant (\textit{i.e.} $h_{rank}$) expectedly performs better in terms of predicting model's bias across categories. The regressor variant (\textit{i.e.} $h_{reg}$) also seems to perform meaningfully, while the classifier (\textit{i.e.} $h_{class}$) variant, as expected does not perform meaningfully, and seems to have a statistically non-significant correlation with high p-values. 

\section{Limitations}
Given that As all of our estimations depend on zero-shot capabilities of a language model, namely CLIP\cite{clip}, disproportionate reliability across different areas of the embedding space could lead to certain semantic concepts being better represented. This shortcoming may be to some extent being modeled by modeling reliability across latent embedding spaces along the lines of \cite{ardeshirembedding}, however, the approach to integrate such notion is non-trivial. We conduct a simple experiment to validate the existence of such effect. We report per-attribute AUROC of clips zero shot accuracy on the test data and report it in Table \ref{tab:per_attr_results}. This is simply measured by performing zero-shot accuracy on an image and passing two textual inputs such as "Young" and "Not Young" and comparing that to the ground-truth value of attribute "Young" for that datapoint. We then measure how this zero-shot accuracy correlates with model robustness across each attribute. We report the results in Table \ref{tab:bias_correlations}. As it can be observed there is a statistically significant correlation. This suggests that on attributes where CLIP prior is stronger, the face-recognition model happens to be more robust. We hypothesize that this may be due to high correlation between the biases of the two models, however, it would be hard to verify this hypothesis.



\begin{table}
\begin{center}
\begin{tabular}{|c|c|c|c|}
\hline
- &  Kendall-tau & Pearson & Spearman \\ \hline
$h_{reg}$ &  (0.19, 3e-04) &  (0.27, 2e-04) &  (0.24, 5e-03)\\ \hline
$h_{class}$ & (0.064, 0.560) & (0.034, 0.834) & (0.072, 0.656) \\ \hline
$h_{rank}$ & \textbf{(0.253, 0.021)} & \textbf{(0.324, 0.040)} & \textbf{(0.366, 0.019)} \\ \hline
\end{tabular}
\caption{Table of correlations between the predicted, and the actual biases. Correlation values are reported as tuples with the format: (correlation-coefficient, p-value). }
\label{tab:correlations}
\end{center}
\end{table}


\begin{table}
\begin{center}
\begin{tabular}{|c|c|c|}
\hline
Kendall-tau & Pearson & Spearman \\ \hline
(0.2743, 0.0126) & (0.291, 0.0679) &  (0.3482, 0.0276)\\ \hline
\end{tabular}
\caption{Bias Correlation. The correlation between the AUROC of a zero-shot clip based classification for a category, compared to the robustness of a face-recognition model across that category.}
\label{tab:bias_correlations}
\end{center}
\end{table}

\begin{table*}
\begin{center}
\begin{tabular}{|c|c|c|c|c|c|}
\hline
attribute &  GT-True AUROC & GT-False AUROC & GT-RMM & CLIP-RMM & 0-shot AUROC\\ \hline
Male & 0.9969 & 0.9912 & 0.0057 & 0.0040 & 0.8932\\
Young & 0.9932 & 0.9939 & 0.0006 & 0.0030 & 0.7501\\
Oval Face & 0.9960 & 0.9923 & 0.0037 & 0.0003 & 0.4775\\
Pale Skin & 0.9921 & 0.9934 & 0.0013 & 0.0010 & 0.7606\\
Big Lips & 0.9918 & 0.9941 & 0.0023 & 0.0016 & 0.5836\\
Big Nose & 0.9961 & 0.9926 & 0.0034 & 0.0003 & 0.5114\\
Pointy Nose & 0.9932 & 0.9934 & 0.0002 & 0.0009 & 0.6190\\
Narrow Eyes & 0.9912 & 0.9937 & 0.0025 & 0.0029 & 0.5345\\
5 o Clock Shadow & 0.9977 & 0.9929 & 0.0047 & 0.0058 & 0.3840\\
Arched Eyebrows & 0.9920 & 0.9939 & 0.0019 & 0.0020 & 0.6638\\
Attractive & 0.9926 & 0.9941 & 0.0014 & 0.0007 & 0.6934\\
Bags Under Eyes & 0.9951 & 0.9929 & 0.0021 & 0.0001 & 0.4412\\
Bald & 0.9957 & 0.9933 & 0.0024 & 0.0049 & 0.5656\\
Bangs & 0.9931 & 0.9934 & 0.0003 & 0.0008 & 0.8513\\
Black Hair & 0.9962 & 0.9923 & 0.0038 & 0.0011 & 0.8187\\
Blond Hair & 0.9898 & 0.9939 & 0.0041 & 0.0039 & 0.8847\\
Blurry & 0.9884 & 0.9936 & 0.0053 & 0.0013 & 0.6912\\
Brown Hair & 0.9926 & 0.9936 & 0.0010 & 0.0023 & 0.7756\\
Bushy Eyebrows & 0.9970 & 0.9928 & 0.0041 & 0.0024 & 0.7114\\
Chubby & 0.9965 & 0.9932 & 0.0032 & 0.0006 & 0.5286\\
Double Chin & 0.9967 & 0.9932 & 0.0034 & 0.0006 & 0.3657\\
Eyeglasses & 0.9936 & 0.9934 & 0.0002 & 0.0041 & 0.8945\\
Goatee & 0.9973 & 0.9932 & 0.0041 & 0.0022 & 0.5699\\
Gray Hair & 0.9972 & 0.9933 & 0.0039 & 0.0030 & 0.7494\\
Heavy Makeup & 0.9931 & 0.9936 & 0.0005 & 0.0015 & 0.4951\\
High Cheekbones & 0.9944 & 0.9923 & 0.0021 & 0.0004 & 0.5354\\
Mouth Slightly Open & 0.9939 & 0.9929 & 0.0009 & 0.0003 & 0.6058\\
Mustache & 0.9972 & 0.9932 & 0.0040 & 0.0021 & 0.5196\\
No Beard & 0.9927 & 0.9974 & 0.0046 & 0.0049 & 0.2825\\
Receding Hairline & 0.9955 & 0.9932 & 0.0023 & 0.0030 & 0.5250\\
Rosy Cheeks & 0.9941 & 0.9933 & 0.0007 & 3.8824 & 0.5282\\
Sideburns & 0.9976 & 0.9932 & 0.0043 & 0.0002 & 0.4113\\
Smiling & 0.9943 & 0.9924 & 0.0018 & 0.0010 & 0.9432\\
Straight Hair & 0.9938 & 0.9933 & 0.0004 & 0.0031 & 0.6136\\
Wavy Hair & 0.9924 & 0.9939 & 0.0015 & 0.0034 & 0.7287\\
Wearing Earrings & 0.9937 & 0.9933 & 0.0004 & 0.0009 & 0.7150\\
Wearing Hat & 0.9937 & 0.9934 & 0.0003 & 0.0012 & 0.8060\\
Wearing Lipstick & 0.9921 & 0.9948 & 0.0026 & 0.0013 & 0.8018\\
Wearing Necklace & 0.9908 & 0.9938 & 0.0029 & 0.0004 & 0.5787\\
Wearing Necktie & 0.9973 & 0.9931 & 0.0041 & 0.0030 & 0.3425\\\hline
\end{tabular}
\caption{per-attribute model performance AUROC, true performance-disparity (GT-RMM) calculated using ground-truth CelebA\cite{celeba} attributes, zero-shot estimated disparity (CLIP-RMM), alongside accuracy of it's zero-shot classification (0-shot AUROC).}
\label{tab:per_attr_results}
\end{center}
\end{table*}

\section{Discussion and Future Work}
Given that this work is in the preliminary stages, there are many aspects of it that could be improved and further explored. As an example, we believe that even though the ranking model seems to lead to better quantitative estimations (according to Table \ref{tab:correlations}), the regressor model has other benefits that may have not been manifested in this experiment. As an example, we believe that since the regression model estimates absolute values of single datapoint loss functions, we can use error bars on the single datapoint loss estimations in conjunction with the pairwise ones to reason about the certainty of the relative predictions. We aim to experiment with different definitions of probability to take that into account. We also believe that not only more extensive experimentation should be conducted across other types of modalities and tasks, but also more extensive experimentation could be done within the confines of the celebA dataset. Given that language models have shown to be better at modeling more specific textual prompts, we suspect that our approach could show its impact more, in more detailed group descriptions (the ones shown in Figure \ref{fig:qualitative}). We aim to explore this in the future.  



\section{Conclusion}
We explored a simple approach to leverage the strengths of language models in zero-shot learning, to provide semantic insights into a trained model's patterns of strengths, weaknesses and biases. Our experiments suggest some promising preliminary results in providing insights in terms a models predicted familiarity and predicted performance on an arbitrary semantic concept. We hope this encourages further investigation in this line of research. 


{\small
\bibliographystyle{ieee_fullname}
\bibliography{egbib}

\begin{thebibliography}{10}\itemsep=-1pt

\bibitem{ardeshirembedding}
Shervin Ardeshir and Navid Azizan.
\newblock Embedding reliability: On the predictability of downstream
  performance.
\newblock In {\em NeurIPS ML Safety Workshop}.

\bibitem{Ardeshir_2022_CVPR}
Shervin Ardeshir, Cristina Segalin, and Nathan Kallus.
\newblock Estimating structural disparities for face models.
\newblock In {\em Proceedings of the IEEE/CVF Conference on Computer Vision and
  Pattern Recognition (CVPR)}, pages 10358--10367, June 2022.

\bibitem{berg2022prompt}
Hugo Berg, Siobhan~Mackenzie Hall, Yash Bhalgat, Wonsuk Yang, Hannah~Rose Kirk,
  Aleksandar Shtedritski, and Max Bain.
\newblock A prompt array keeps the bias away: Debiasing vision-language models
  with adversarial learning.
\newblock {\em arXiv preprint arXiv:2203.11933}, 2022.

\bibitem{gpt3}
Tom Brown, Benjamin Mann, Nick Ryder, Melanie Subbiah, Jared~D Kaplan, Prafulla
  Dhariwal, Arvind Neelakantan, Pranav Shyam, Girish Sastry, Amanda Askell,
  et~al.
\newblock Language models are few-shot learners.
\newblock {\em Advances in neural information processing systems},
  33:1877--1901, 2020.

\bibitem{chowdhery2022palm}
Aakanksha Chowdhery, Sharan Narang, Jacob Devlin, Maarten Bosma, Gaurav Mishra,
  Adam Roberts, Paul Barham, Hyung~Won Chung, Charles Sutton, Sebastian
  Gehrmann, et~al.
\newblock Palm: Scaling language modeling with pathways.
\newblock {\em arXiv preprint arXiv:2204.02311}, 2022.

\bibitem{harkonen2020ganspace}
Erik H{\"a}rk{\"o}nen, Aaron Hertzmann, Jaakko Lehtinen, and Sylvain Paris.
\newblock Ganspace: Discovering interpretable gan controls.
\newblock {\em Advances in Neural Information Processing Systems},
  33:9841--9850, 2020.

\bibitem{hong2022avatarclip}
Fangzhou Hong, Mingyuan Zhang, Liang Pan, Zhongang Cai, Lei Yang, and Ziwei
  Liu.
\newblock Avatarclip: Zero-shot text-driven generation and animation of 3d
  avatars.
\newblock {\em arXiv preprint arXiv:2205.08535}, 2022.

\bibitem{joshi2019semantic}
Ameya Joshi, Amitangshu Mukherjee, Soumik Sarkar, and Chinmay Hegde.
\newblock Semantic adversarial attacks: Parametric transformations that fool
  deep classifiers.
\newblock In {\em Proceedings of the IEEE/CVF international conference on
  computer vision}, pages 4773--4783, 2019.

\bibitem{cifar}
Alex Krizhevsky, Geoffrey Hinton, et~al.
\newblock Learning multiple layers of features from tiny images.
\newblock 2009.

\bibitem{celeba}
Ziwei Liu, Ping Luo, Xiaogang Wang, and Xiaoou Tang.
\newblock Deep learning face attributes in the wild.
\newblock In {\em Proceedings of International Conference on Computer Vision
  (ICCV)}, December 2015.

\bibitem{qi2023improving}
Qi Qi and Shervin Ardeshir.
\newblock Improving identity-robustness for face models.
\newblock {\em arXiv preprint arXiv:2304.03838}, 2023.

\bibitem{qiu2020semanticadv}
Haonan Qiu, Chaowei Xiao, Lei Yang, Xinchen Yan, Honglak Lee, and Bo Li.
\newblock Semanticadv: Generating adversarial examples via
  attribute-conditioned image editing.
\newblock In {\em Computer Vision--ECCV 2020: 16th European Conference,
  Glasgow, UK, August 23--28, 2020, Proceedings, Part XIV 16}, pages 19--37.
  Springer, 2020.

\bibitem{clip}
Alec Radford, Jong~Wook Kim, Chris Hallacy, Aditya Ramesh, Gabriel Goh,
  Sandhini Agarwal, Girish Sastry, Amanda Askell, Pamela Mishkin, Jack Clark,
  et~al.
\newblock Learning transferable visual models from natural language
  supervision.
\newblock In {\em International conference on machine learning}, pages
  8748--8763. PMLR, 2021.

\bibitem{ramaswamy2021fair}
Vikram~V Ramaswamy, Sunnie~SY Kim, and Olga Russakovsky.
\newblock Fair attribute classification through latent space de-biasing.
\newblock In {\em Proceedings of the IEEE/CVF conference on computer vision and
  pattern recognition}, pages 9301--9310, 2021.

\bibitem{rawls2001justice}
John Rawls.
\newblock {\em Justice as fairness: A restatement}.
\newblock Harvard University Press, 2001.

\bibitem{sanghi2022clip}
Aditya Sanghi, Hang Chu, Joseph~G Lambourne, Ye Wang, Chin-Yi Cheng, Marco
  Fumero, and Kamal~Rahimi Malekshan.
\newblock Clip-forge: Towards zero-shot text-to-shape generation.
\newblock In {\em Proceedings of the IEEE/CVF Conference on Computer Vision and
  Pattern Recognition}, pages 18603--18613, 2022.

\bibitem{sauer2021counterfactual}
Axel Sauer and Andreas Geiger.
\newblock Counterfactual generative networks.
\newblock {\em arXiv preprint arXiv:2101.06046}, 2021.

\bibitem{touvron2023llama}
Hugo Touvron, Thibaut Lavril, Gautier Izacard, Xavier Martinet, Marie-Anne
  Lachaux, Timoth{\'e}e Lacroix, Baptiste Rozi{\`e}re, Naman Goyal, Eric
  Hambro, Faisal Azhar, et~al.
\newblock Llama: Open and efficient foundation language models.
\newblock {\em arXiv preprint arXiv:2302.13971}, 2023.

\bibitem{wang2022clip}
Can Wang, Menglei Chai, Mingming He, Dongdong Chen, and Jing Liao.
\newblock Clip-nerf: Text-and-image driven manipulation of neural radiance
  fields.
\newblock In {\em Proceedings of the IEEE/CVF Conference on Computer Vision and
  Pattern Recognition}, pages 3835--3844, 2022.

\end{thebibliography}
}

\end{document}